\documentclass{article}

\usepackage{amsmath}
\usepackage{times}
\usepackage{graphicx} % more modern
\usepackage{subfigure} 
\usepackage{natbib}
\usepackage{microtype}
\usepackage{algorithm}
\usepackage{algorithmic}

\usepackage{hyperref}

\usepackage{amsbsy}
\usepackage[accepted]{icml2016} 

\newcommand{\vect}[1]{\mathbf{#1}}
\newcommand{\vects}[1]{\boldsymbol{#1}}

\newcommand{\vb}[0]{\vect{b}}

\newcommand{\vh}[0]{\vect{h}}

\newcommand{\vx}[0]{\vect{x}}
\newcommand{\vy}[0]{\vect{y}}

\newcommand{\params}[0]{\vects{\theta}}
\newcommand{\hypers}[0]{\vects{\lambda}}

\newcommand{\vw}[0]{\vect{W}}

\icmltitlerunning{Scalable Gradient-Based Tuning of Continuous Regularization Hyperparameters}

\begin{document} 

\twocolumn[
\icmltitle{Scalable Gradient-Based Tuning of \\ Continuous Regularization Hyperparameters}

\icmlauthor{Jelena Luketina\texorpdfstring{$^1$}{1}}{jelena.luketina@aalto.fi}
\icmlauthor{Mathias Berglund\texorpdfstring{$^1$}{1}}{mathias.berglund@aalto.fi}
\icmlauthor{Klaus Greff\texorpdfstring{$^2$}{2}}{klaus@idsia.ch}
\icmlauthor{Tapani Raiko\texorpdfstring{$^1$}{1}}{tapani.raiko@aalto.fi}
\icmladdress{\texorpdfstring{$^1$}{1}Department of Computer Science, Aalto University, Finland}
\icmladdress{\texorpdfstring{$^2$}{2}IDSIA, Dalle Molle Institute for Artificial Intelligence, USI-SUPSI, Manno-Lugano, Switzerland}

\icmlkeywords{hyperparameter optimization, gradient-based hyperparameter selection, deep learning, ICML}

\vskip 0.3in
]

\begin{abstract} 
Hyperparameter selection generally relies on running multiple full training trials, with selection based on validation set performance. We propose a gradient-based approach for locally adjusting hyperparameters during training of the model. Hyperparameters are adjusted so as to make the model parameter gradients, and hence updates, more advantageous for the validation cost. We explore the approach for tuning regularization hyperparameters and find that in experiments on MNIST, SVHN and CIFAR-10, the resulting regularization levels are within the optimal regions. The additional computational cost depends on how frequently the hyperparameters are trained, but the tested scheme adds only 30\% computational overhead regardless of the model size. Since the method is significantly less computationally demanding compared to similar gradient-based approaches to hyperparameter optimization, and consistently finds good hyperparameter values, it can be a useful tool for training neural network models.
\end{abstract} 

% % % % % % % % % % % % % % 
\section{Introduction}
% % % % % % % % % % % % % % 

Specifying and training artificial neural networks requires several design choices that are often not trivial to make. Many of these design choices boil down to the selection of hyperparameters. The process of hyperparameter selection is in practice often based on trial-and-error and grid or random search \citep{bergstra2012random}. There are also a number of automated methods \citep{bergstra2011hyperparam,snoek2012bayesian}, all of which rely on multiple complete training runs with varied fixed hyperparameters, with the hyperparameter selection based on the validation set performance.

Although effective, these methods are expensive as the user needs to run multiple full training runs. In the worst case, the number of needed runs also increases exponentially with the number of hyperparameters to tune, if an extensive exploration is desired. In many practical applications such an approach is too tedious and time-consuming, and it would be useful if a method existed that could automatically find acceptable hyperparameter values in one training run even if the user did not have a strong intuition regarding good values to try for the hyperparameters.

In contrast to these methods, we treat hyperparameters similar to elementary\footnote{Borrowing the expression from \citet{maclaurin2015gradient}, we refer to the model parameters customarily trained with back-propagation as \textit{elementary} parameters, and to all other parameters as hyperparameters.} parameters during training, in that we simultaneously update both sets of parameters using stochastic gradient descent. The gradient of elementary parameters is computed as in usual training from the cost of the regularized model on the training set, while the gradient of hyperparameters (hypergradient) comes from the cost of the unregularized model on the validation set. For simplicity, we will refer to the training set as $T_1$ and  to the validation set (or any other data set used exclusively for training the hyperparameters) as $T_2$. The method itself will be called $T_1-T_2$, referring to the two simultaneous optimization processes. 

Similar approaches have been proposed since the late 1990s; however, these methods either require computation of the inverse Hessian \citep{larsen1998adaptive, bengio2000gradient, chen99optimal, ng08efficient}, or propagating gradients through the entire history of parameter updates \citet{maclaurin2015gradient}. Moreover, these methods make changes to the hyperparameter only once the elementary parameter training has ended. These drawbacks make them too expensive for use in modern neural networks, which often require millions of parameters and large data sets.

% - Relevance of regularization:
% how tedious it is to deal with, examples of SOTA (and other) models utilizing additive noise and/or L2-type regularizers (?). 
% Examples of papers where a better choice of hyperparameters or richer hyper-parametrization could lead to better results.  ()

\begin{figure*}[htb]
\begin{flushright}
\centerline{\includegraphics[width=1\textwidth]{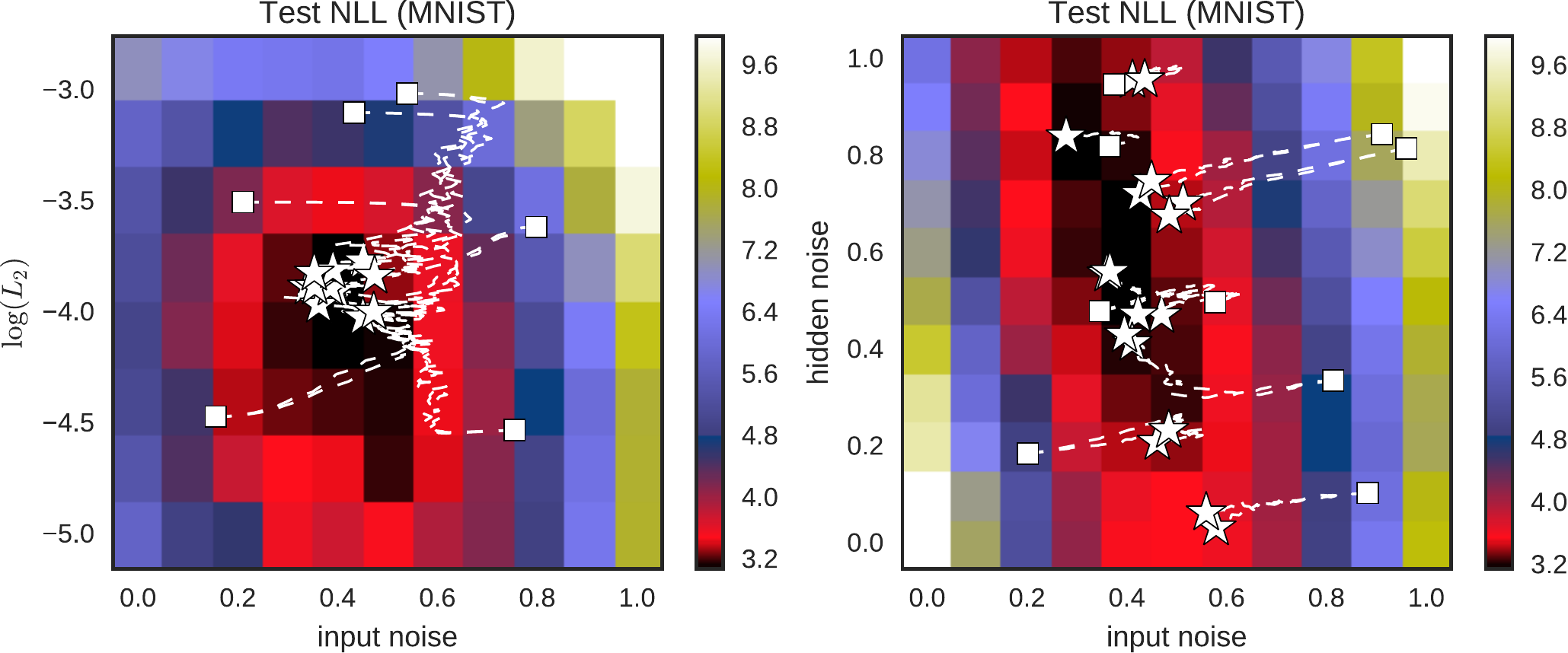}}
\caption{Left: Values of additive input noise and $L_2$ penalty $(n_0, log(l_2))$ during training using the $T_1-T_2$ method for hyperparameter tuning. Trajectories are plotted over the grid search result for the same regularization pair. Initial hyperparameter values are denoted with a square, final hyperparameter values are denoted with a star. Right: Similarly constructed trajectories, on a model regularized with input and hidden layer additive noise $(n_0, n_1)$.}
\label{fig:traject}
\end{flushright}
\end{figure*}

Elements distinguishing our approach are:\vspace{-1em}
\begin{enumerate}
\item  By making some very rough approximations, our method for modifying hyperparameters avoids using computationally expensive terms, including the computation of the Hessian or its inverse. 
This is because with the $T_1-T_2$ method, hyperparameter updates are based on stochastic gradient descent, instead of Newton's method. Furthermore, any dependency of elementary parameters on hyperparameters beyond the last update is disregarded. As a result, additional computational and memory overhead therefore becomes comparable to back-propagation.    
\item  Hyperparameters are trained simultaneously with elementary parameters. Feedback and feedforward passes can be computed simultaneously for the training and validation set, further reducing the computational cost.
\item  We add batch normalization \citep{ioffe2015batchnorm} and adaptive learning rates \citep{kingma2015adam} to the process of hyperparameter training, which diminishes some of the problems of gradient-based hyperparameter optimization. 
Through batch normalization, we can counter internal covariate shifts. This eliminates the need for different learning rates at each layer, as well as speeding up adjustment of the elementary parameters to the changes in hyperparameters. This is particularly relevant when parametrizing each of the layers with a separate hyperparameter. 

\end{enumerate}

A common assumption is that the choice of hyperparameters affects the whole training trajectory, i.e. changing a hyperparameter on the fly during training has a significant effect on the training trajectory. This ``hysteresis effect'' implies that in order to measure how a hyperparameter combination influences the validation set performance, the hyperparameters need to be kept fixed during the whole training procedure. However, to our knowledge this has not been systematically studied. If the hysteresis effect is weak enough and the largest changes to the hyperparameter happen early on, it becomes possible to train the model while tuning the hyperparameters on the fly during training, and then use the final hyperparameter values to retrain the model if a fixed set of hyperparameters is desired. We also explore this approach.

An important design choice when training neural network models is which regularization strategy to use in order to ensure that the model generalizes to data not included in the training set. Common regularization strategies involve adding explicit terms to the model or the cost function during training, such as penalty terms on the model weights or injecting noise to inputs or neuron activations. Injecting noise is particularly relevant for denoising autoencoders and related models \citep{vincent10stacked, rasmus2015ladder}, where performance strongly depends on the level of noise.

% - How we test the method, why do we test regularizers and not other continuous hyperparameters?
% Well, that is a good question. Intuition is that regularizers might be easier to train using simultaneous updates of elementary parameters and hyperparameters, due to there existing little long term dependencies (unlike with learning rate and momentum schedules); while the validation loss depends smoothly on the hyperparameter value (as example, see grid search result for additive noise\ref{fig:traject}). 

Although the proposed method could work in principle for any continuous hyperparameter, we have specifically focused on studying tuning of regularization hyperparameters. We have chosen to use Gaussian noise added to the inputs and hidden layer activations, in addition to $L_2$ weight penalty. A third, often used, regularization method that involves a hyperparameter choice is dropout \citep{srivastava2014dropout}. However, we have omitted studying dropout as it is not trivial to compute a gradient on the dropout rate. Moreover, dropout can be seen as a form of multiplicative Gaussian noise \citep{wang2013fastdropout}. We also omit study adapting the learning rate, since we suspect that the local gradient information is not sufficient to determine optimal learning rates.

In Section~\ref{sec:model} we present details on the proposed method. The method is tested with multiple MLP and CNN network structures and regularization schemes, detailed in Section~\ref{sec:experiments}. The results of the experiments are presented in Section~\ref{sec:results}.

% % % % % % % % % % % % % % 
\section{Proposed Method}
% % % % % % % % % % % % % % 
\label{sec:model}

We propose a method, $T_1-T_2$, for tuning continuous hyperparameters of a model using the gradient of the performance of the model on a separate validation set $T_2$. In essence, we train a neural network model on a training set $T_1$ as usual. However, for each update of the network weights and biases, i.e. the elementary parameters of the network, we tune the hyperparameters so as to make the direction of the weight update as beneficial as possible for the validation cost on a separate dataset $T_2$.

Formally, when training a neural network model, we try to minimize an objective function that depends on the training set, model weights and hyperparameters that determine the strength of possible regularization terms. When using gradient descent, we denote the optimization objective function $\tilde{C_1}(\cdot)$ and the corresponding weight update as:
\begin{align}  
\tilde{C_1}(\params | \hypers,  T_1) &= C_1(\params | \hypers,  T_1) + \Omega(\params,\hypers) , \\
  {\params}_{t+1}&={\params}_t + \eta_1 \nabla_{\params} \tilde{C_1} ({\params}_t | {\hypers}_t,  T_1),
  \label{eq:W_update}
\end{align}

where $\tilde{C_1}(\cdot)$  and $\Omega(\cdot)$ are cost and regularization penalty terms, $T_1=\{(\vx_{i},\ \vy_{i})\}$  is the training data set, $\params=\{\vw^{l},\ \vb^{l}\}$ a set of elementary parameters including weights and biases of each layer, $\hypers$ denotes various hyperparameters that determine the strength of regularization, while $\eta_1$ is a learning rate. Subscript $t$ refers to the iteration number.

Assuming $T_2=\{(\vx_{i},\ \vy_{i})\}$ is a separate validation data set, the generalization performance of the model is measured with a validation cost $C_2(\params_{t+1}, T_2)$, which is usually a function of the unregularized model. Hence the value of the cost function of the actual performance of the model does not depend on the regularizer directly, but only on the elementary parameter updates. The gradient of the validation cost with respect to $ \hypers$ is:
    \[ \nabla_{ \hypers} C_2=  (\nabla_{ \params}
    C_2) ( \nabla_{ \hypers} \params_{t+1})\]
We only consider the influence of the regularization hyperparameter on the current elementary parameter update, $ \nabla_{ \hypers} \params_{t+1}  =  \eta_1 \nabla_{ \hypers} \nabla_{ \params} \tilde{C_1}$ based on Eq.\ (\ref{eq:W_update}). The hyperparameter update is therefore: 
\begin{align} 
\hypers_{t+1} = \hypers_t + \eta_2 (\nabla_{ \params} C_2)  (\nabla_{ \hypers} \nabla_{ \params} \tilde{C_1})
\label{eq:lambdaupdate}
\end{align}
where $\eta_2$ is a learning rate.

The method is greedy in the sense that it only depends on one parameter update, and hence rests on the assumption that a good hyperparameter choice can be evaluated based on the local information within only one elementary parameter update.

\subsection{Motivation and analysis}

The most similar previously proposed model is the incremental gradient version of the hyperparameter update  from \citep{chen99optimal}. However their derivation of the hypergradient assumes a Gauss-Newton update of the elementary parameters, making computation of the gradient and the hypergradient significantly more expensive. 

A well justified closed form for the term $\nabla_{ \hypers} \params$  is available once the elementary gradient has converged \citep{ng08efficient}, with the update of the form (\ref{eq:ngupdate}). Comparing this expression with the $T_1-T_2$ update, (\ref{eq:lambdaupdate}) can be considered as approximating (\ref{eq:ngupdate}) in the case when gradient is near convergence and the Hessian can be well approximated by identity $\nabla_{ \params}^2 \tilde{C_1} = I$: 
\begin{align} 
\hypers_{t+1} &= \hypers_t +  
(\nabla_{ \params} C_2)  (\nabla_{ \params}^{2}\tilde{C_1})^{-1} (\nabla_{ \hypers} \nabla_{ \params} \tilde{C_1}).
\label{eq:ngupdate}
\end{align}

\begin{figure}[!t]
\centering
\includegraphics[width=1\columnwidth]{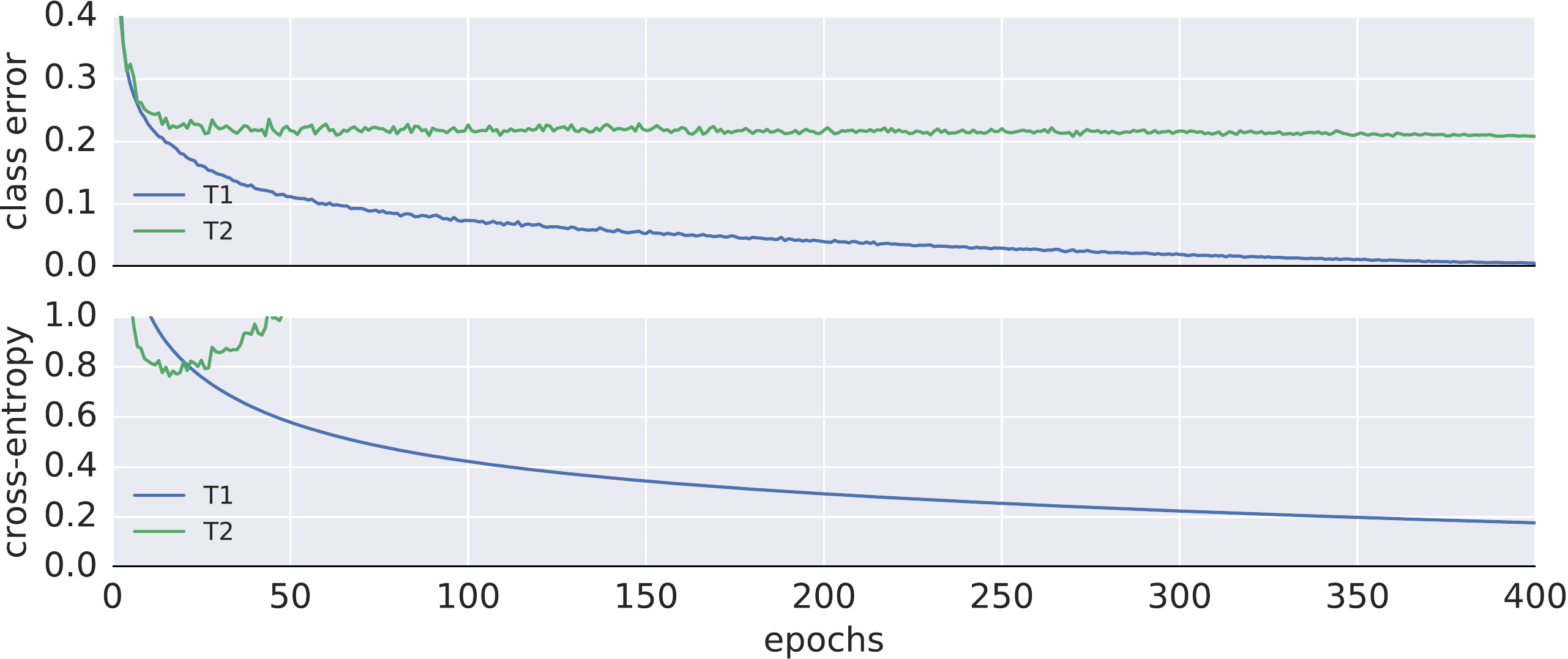}
\includegraphics[width=1\columnwidth]{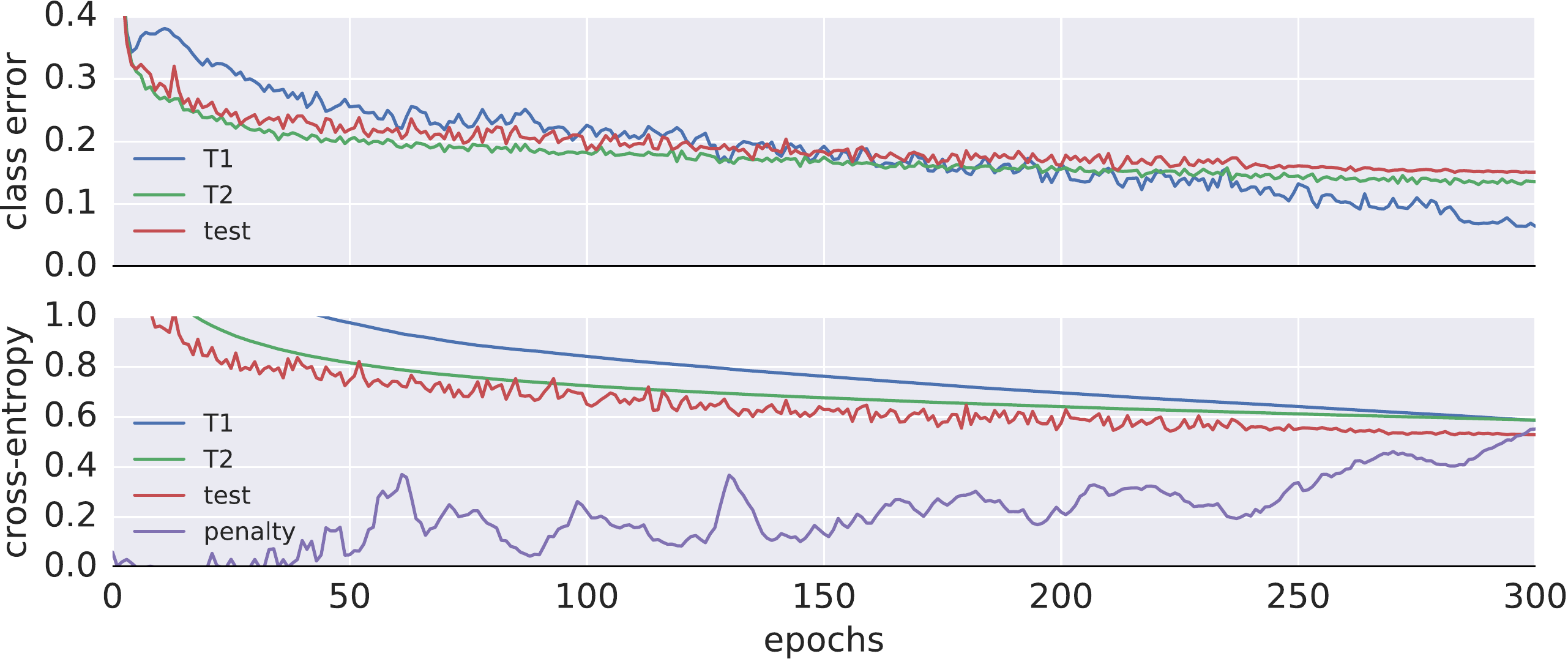}
\caption{Evolution of classification error and cross-entropy over the course of training, for a single SVHN experiment. Top: evolution of the classification error and costs with a fixed choice of hyperparameters $(n_0, \mathbf{l_2})$. Bottom: classification error and costs during training with $T_1-T_2$, using $(n_0, \mathbf{l_2})$ as initial hyperparameter values. $T_1-T_2$ prevents otherwise strong overfitting.}
\label{fig:errorcurves}
\end{figure}

Another approach to hypergradient computation is given in \citet{maclaurin2015gradient}. There, the term $\nabla_{ \hypers} \params_T$ ($T$ denoting the final iteration number) considers effect of the hyperparameter on the entire history of updates:
\begin{align} 
\params_T &= \params_{0}+ \sum_{0<k<T}\triangle \params_{k,k+1}(\params_{k}(\hypers_t),\ \hypers_t,\ \eta_{k}),
\\
\hypers_{t+1} &= \hypers_t +  
(\nabla_{ \params} C_2) (\nabla_{ \hypers} \params_{T}). 
\end{align}

In the simplest case where the update is formed only with the current gradient $\triangle \params_{k,k+1}=-\eta_{1,k} \nabla_{\params}\tilde{C}_1$, i.e. not including the momentum term or adaptive learning rates, the update of a hyperparameter is formed by collecting the elements from the entire training procedure:
\begin{align}
\hypers_{t+1} = \hypers_t + \eta_{2} \nabla_{\params} C_{2} \sum_{0<k<T} \eta_1 \nabla_{\hypers}\nabla_{\params} \tilde{C}_{1,k}. 
\label{eq:mclaurinupdate2}
\end{align} 

Eq.\ (\ref{eq:lambdaupdate}) can therefore be considered as an approximation of (\ref{eq:mclaurinupdate2}), where we consider only the last update instead of backpropagating through the whole weight update history and updating the hyperparameters without resetting the weights.

In theory, approximating the Hessian with identity might cause difficulties.
From Equation (\ref{eq:lambdaupdate}), it follows that the method converges when $(\nabla_{ \params} C_2)  (\nabla_{ \hypers} \nabla_{ \params} \tilde{C}_1) = \vects{0}$, or in other words, for all components $i$ of the hyperparameter vector $\hypers$, $\nabla_{ \params} C_2$ is orthogonal to $\frac{\partial \nabla_{ \params} \tilde{C}_1}{\partial \lambda_i}$. This is in contrast to the standard optimization processes that converge when the gradient is zero. In fact, we cannot guarantee convergence at all. Furthermore, if we replace the global (scalar) learning rate $\eta_1$ in Equation (\ref{eq:W_update}) with individual learning rates $\eta_{1,j}$ for each elementary-parameter $\theta_{j,t}$, the point of convergence could change.

It is clear that the identity Hessian assumption is an approximation that will not hold strictly. However, arguably, batch normalization  \citep{ioffe2015batchnorm} is eliminating part of the problem, by making the Hessian closer to identity \citep{vatanen2013pushing,raiko2012deep}, making the approximation more justified. Another step towards making even closer approximation are transformations that further whiten the hidden representations \citep{desjardins2015natnet}. 

\begin{figure}[!t]
\centering
\includegraphics[width=1\columnwidth]{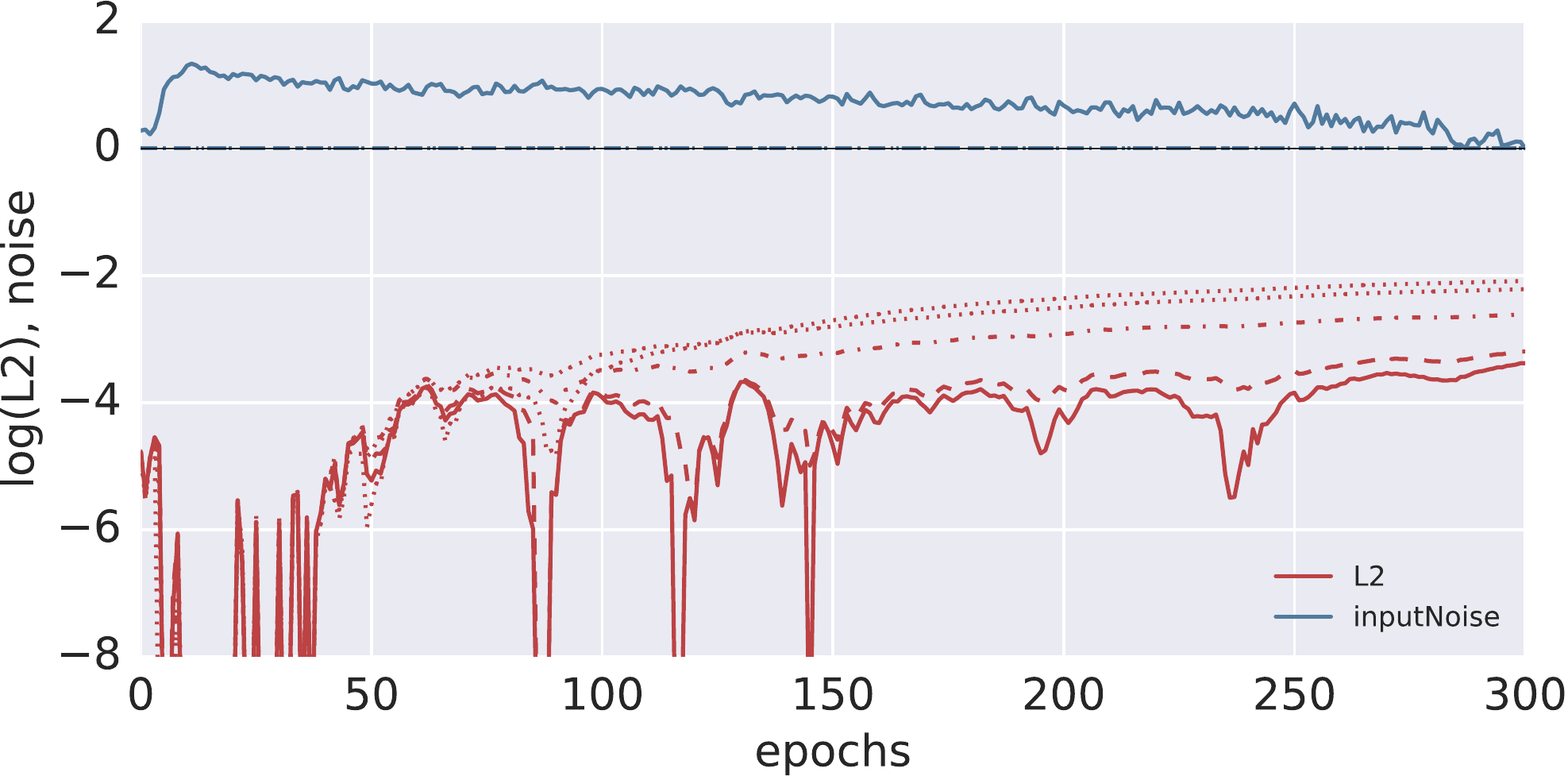}
\caption{Evolution of the hyperparameters for the same SVHN experiment as in \autoref{fig:errorcurves} bottom. The green curve at the top shows the standard deviation of the input noise. The red curves at the bottom demonstrate values of $L2$ regularization of each layer on a log scale. The solid line refers to the input layer and least solid one to the top layer. The Figure illustrates two common patterns we observed for SVHN: firstly the noise level tends to increase in the beginning and decrease later during training, and secondly the $L2$ decay always ends up stronger for the higher layers.}
\label{fig:hyperevolve}
\end{figure}

\subsection{Computational cost}

The most computationally expensive term of the proposed method is $(\nabla_{ \params} C_2)  (\nabla_{ \hypers} \nabla_{ \params} \tilde{C}_1)$, where the exact complexity depends on the details of the implementation and the hyperparameter. When training $L2$ penalty regularizers $\Omega(\params)=\sum_j \frac{\lambda_j}{2} \theta_j^2$, the additional cost is negligible, as $\frac{\partial^2 \tilde{C}_1}{\partial \lambda_j \partial \theta_k }=\theta_j \delta_{j,k}$, where $\delta$ is the indicator function.

The gradient of the cost with respect to a noise hyperparameter $\sigma_l$ at layer $l$, can be computed as $\frac{\partial \tilde{C}_1}{\partial \sigma_l} = (\nabla_{ \vh_l} \tilde{C}_1)\left(\frac{\partial \vh_l}{\partial \sigma_l}\right)^\top$, where $\vh_l$ is hidden layer $l$ activation. In case of additive Gaussian noise, where noise is added as $\vh_l \rightarrow \vh_l + \sigma_L \vect{e}$, where $\vect{e}$ is a random vector sampled from the standard normal distribution with the same dimensionality as $\vh_l$, the derivative becomes $\frac{\partial \tilde{C}_1}{\partial \sigma_l} = \frac{\partial \tilde{C}_1}{\partial \vh_l} \vect{e}^\top$. It can be shown that the cost of computing this term scales comparably to backpropagation, due to the properties of R and L-operators \citep{pearlmutter1994hessian, schraudolph2002}.

For our implementation, the cost of computing the hypergradients of a model with additive noise in each layer, was around 3 times that of backpropagation. We reduced this cost further by making one hyperparameter update per each 10 elementary parameter updates. While it did not change performance of the method, it reduced the additional cost to about only 30\% that of backpropagation. The cost could be possibly reduced even further by making hyperparameter updates even less frequently, though we have not explored this further.

\begin{figure*}[htb]
\centering
\includegraphics[width=0.32\textwidth]{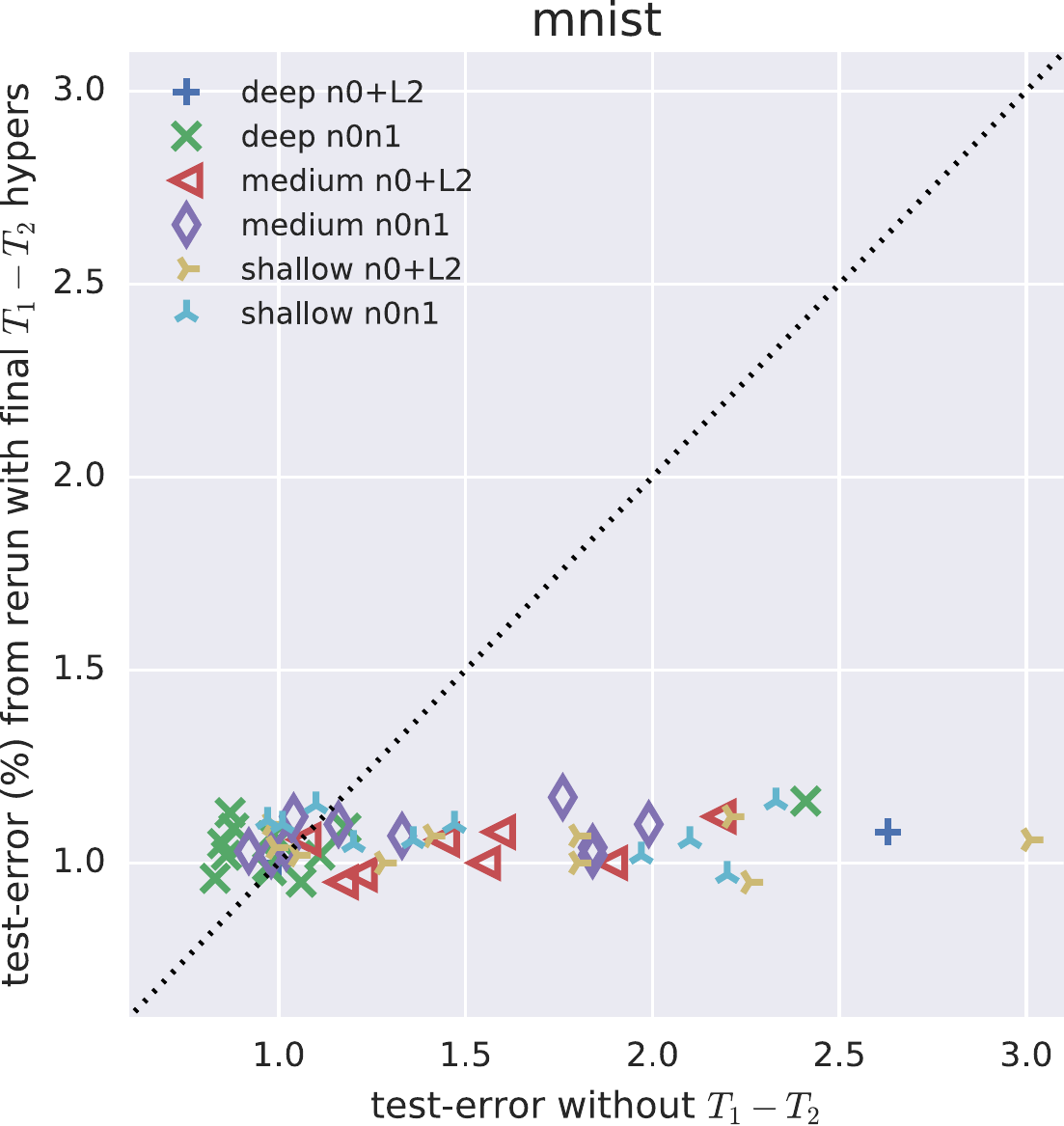}
\hfill
\includegraphics[width=0.32\textwidth]{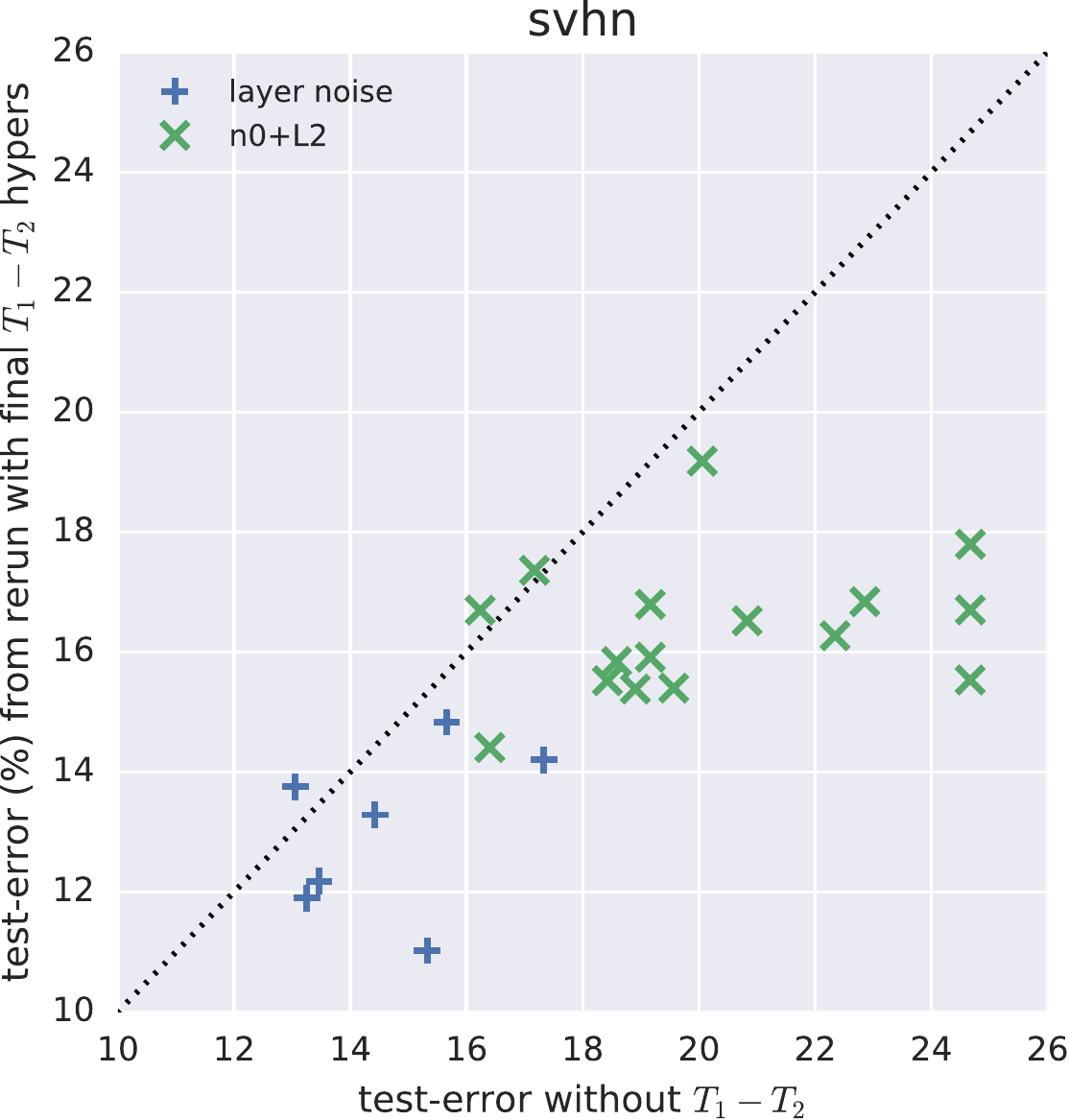}
\hfill
\includegraphics[width=0.32\textwidth]{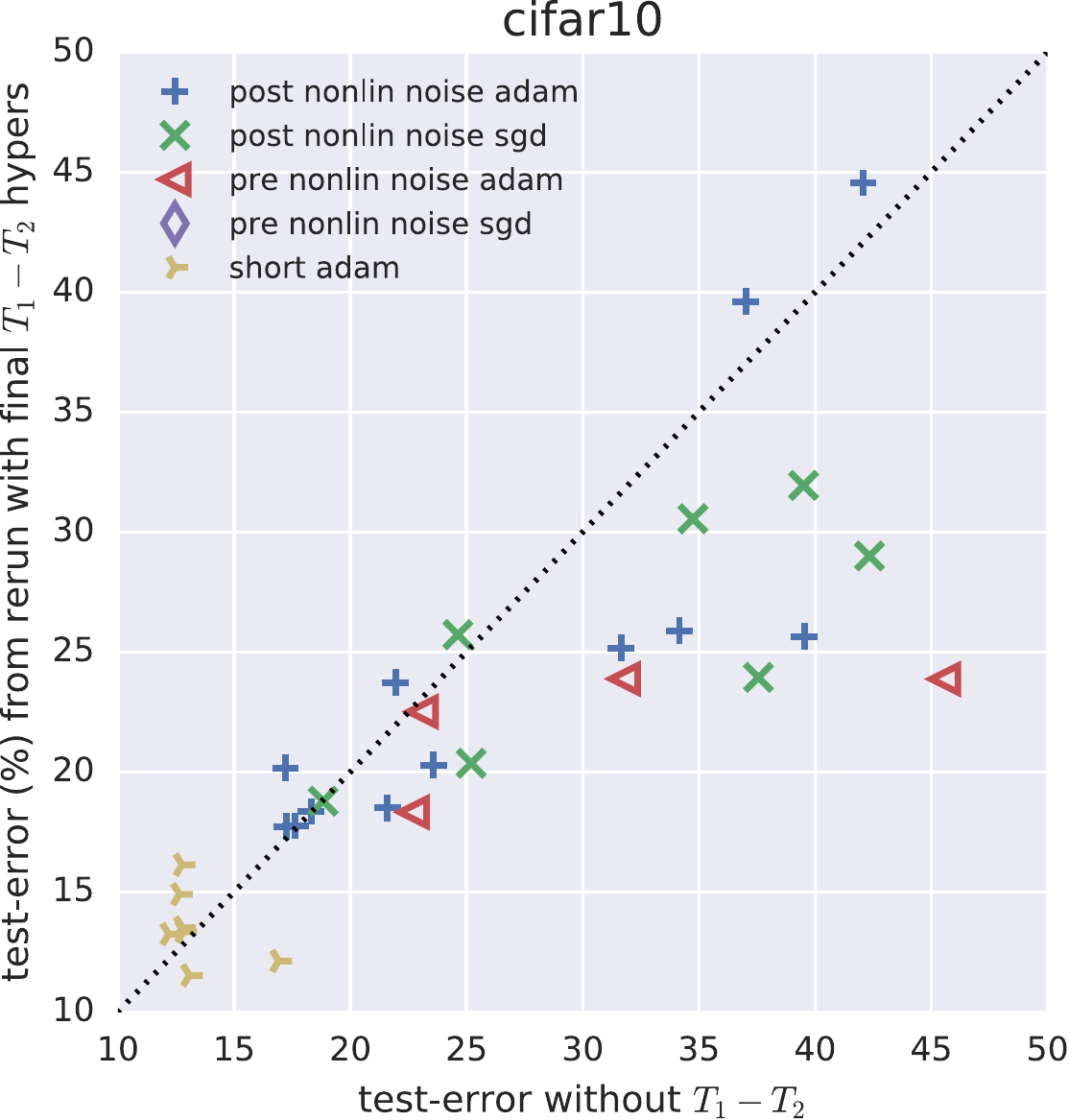}
\caption{Comparison of test performances when training with fixed hyperparameters before and after tuning them with $T_1-T_2$. Results for MNIST are shown on the left, SVHN in the middle, CIFAR-10 on the right. Points are generated on a variety of network configurations, where equal symbols mark equal setup.}
\label{fig:before}
\end{figure*}

% % % % % % % % % % % % % % 
\section{Experiments}
% % % % % % % % % % % % % % 
\label{sec:experiments}

The goal of the experimental section is to address the following questions: 
\pagebreak
\begin{itemize}
\item Will the method find new hyperparameters which improve the performance of the model, compared to the initial set of hyperparameters?
\item Can we observe hysteresis effects, i.e. will the model obtained, while simultaneously modifying parameters and hyperparameters, perform the same as a model trained with a hyperparameter fixed to the final value?
\item Can we observe overfitting on the validation set $T_2$? When hyperparameters are tuned for validation performance, is the performance on the validation set still indicative of the performance on the test set? 
\end{itemize}

We test the method on various configurations of multilayer perceptrons (MLPs) with ReLU activation functions \citep{relu2013} trained on the MNIST \citep{mnist} and SVHN \citep{svhn} data set. We also test the method on two convolutional architectures (CNNs) using CIFAR-10 \citep{cifar10}. The CNN architectures used were modified versions of model All-CNN-C from \citep{springenberg_striving_2014} and a baseline model from \citep{rasmus2015ladder}, using ReLU and leakyReLU activations \citep{leaky_relu}. The models were implemented with the Theano package \citep{the_theano_development_team_theano:_2016}. All the code, as well as the exact configurations used in the experiments can be found in the project's Github repository\footnote{https://github.com/jelennal/t1t2}. 

For MNIST we tried various network sizes: shallow $1000 \times 1000 \times 1000$ to deep $4000 \times 2000 \times 1000 \times 500 \times 250$. Training set $T_1$ had 55~000 samples, and validation $T_2$ had 5~000 samples. The split between $T_1$ and $T_2$ was made using a different random seed in each of the experiments to avoid overfitting to a particular subset of the training set. Data preprocessing consisted of only centering each feature.

In experiments with SVHN we tried $2000 \times 2000 \times 2000$ and  $4000 \times 2000 \times 1000 \times 500 \times 250$ architectures. Global contrast normalization was used as the only preprocessing step. Out of 73257 training samples, we picked a random 65 000 samples for $T_1$ and the remaining 8 257 samples for $T_2$. None of the SVHN experiments used tied hyperparameters, i.e. each layer was parametrized with a separate hyperparameter, which was tuned independently.     

To test on CIFAR-10 with convolutional networks, we used 45 000 samples for $T_1$ and 5 000 samples for $T_2$. The data was preprocessed using global contrast normalization and whitening. 

\begin{figure*}
\centering
\includegraphics[width=0.32\textwidth]{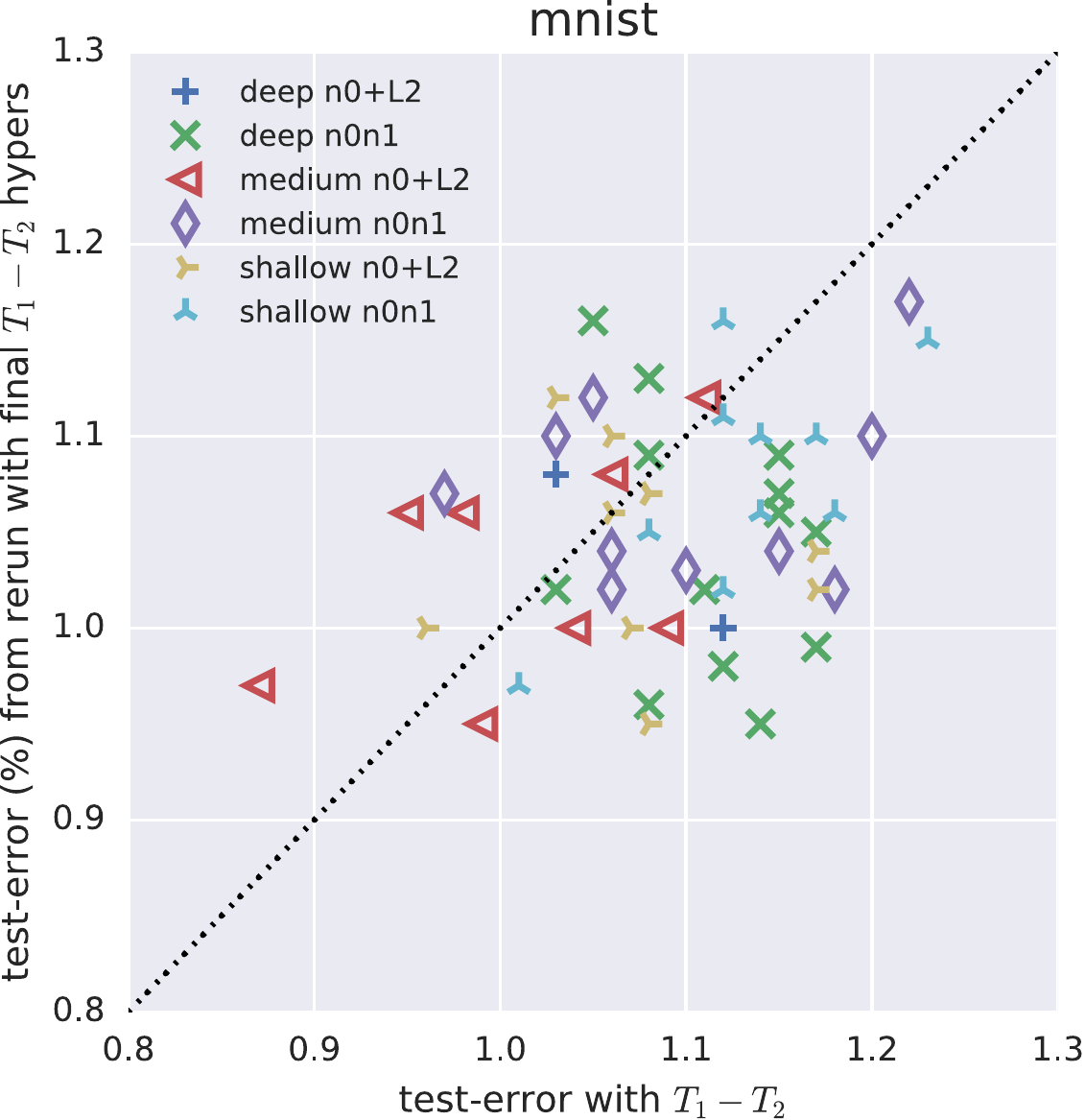}
\hfill
\includegraphics[width=0.32\textwidth]{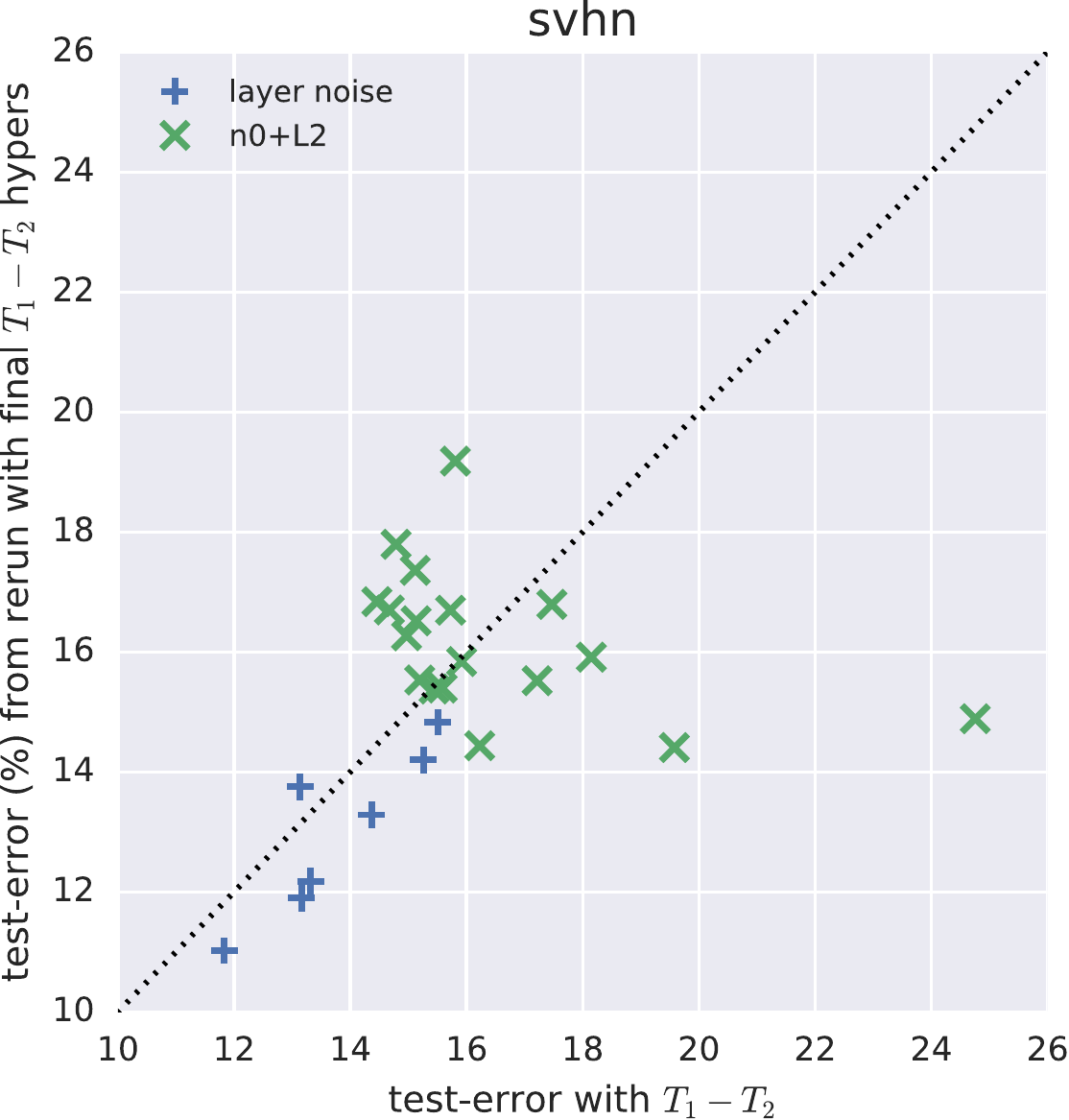}
\hfill
\includegraphics[width=0.32\textwidth]{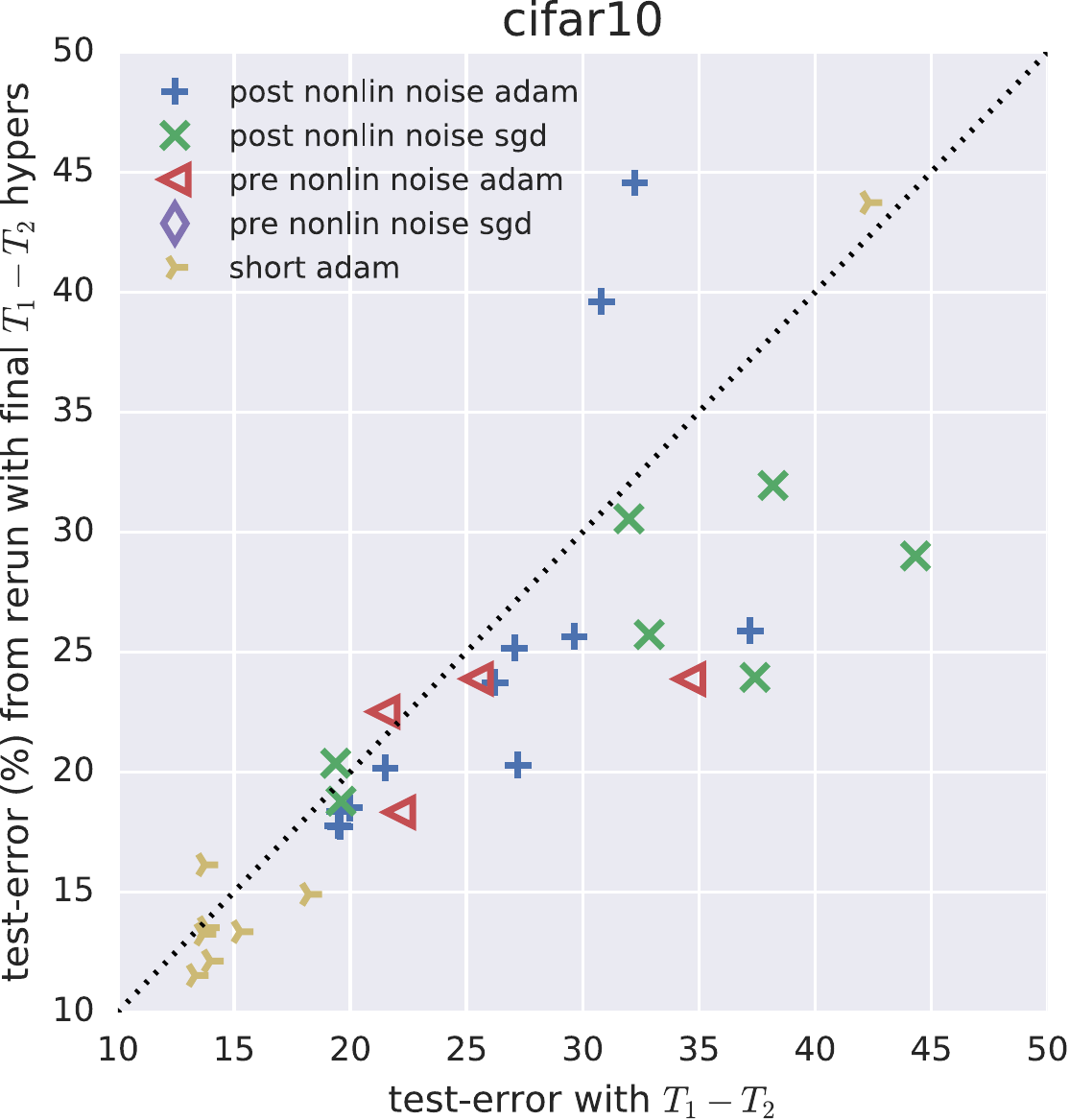}
\caption{Test error after one run with $T_1-T_2$ compared to a rerun where we use the final values of the hyperparameters at the end of $T_1-T_2$ training as fixed hyperparameters for a new run (left: MNIST, middle: SVHN, right: CIFAR-10). Th correlation indicates that $T_1-T_2$ is useful also for finding approximate hyperparameters for training without an adaptive hyperparameter method.}
\label{fig:q2}
\end{figure*}

We regularized the models with additive Gaussian noise to the input with standard deviation $n_0$ and each hidden layer with standard deviation $n_1$; or a combination of additive noise to the input layer and L2 penalty with strength multiplier $l_2$ for weights in each of the layers. Because L2 penalty matters less in models using batch normalization, in experiments using L2 penalty we did not use batch normalization. We tried both tied regularization levels (one hyperparameter for all hidden layers) and having separate regularization parameters for each layer. As a cost function, we use cross-entropy for both $T_1$ and $T_2$.

Each of the experiments were run with 200-300 epochs, using batch size 100 for both elementary and hyperparameter training. To speed up elementary parameter training, we use an annealed ADAM learning rate schedule \citep{kingma2015adam} with a step size of $10^{-3}$ (MLPs) or $2\cdot 10^{-3}$ (CNNs). For tuning noise hyperparameters, we use vanilla gradient descent with a step size $10^{-1}$; while for L2 hyperparameters, step sizes were significantly smaller, $10^{-4}$. In experiments on larger networks we also use ADAM for tuning hyperparameters, with the step size $10^{-3}$ for noise and $10^{-6}$ for L2. We found that while the learning rate did not significantly influence the general area of convergence for a hyperparameter, too high learning rates did cause too noisy and sudden hyperparameter changes, while too low learning rates resulted in no significant changes of hyperparameters. A rule of thumb is to use a learning rate corresponding to the expected order of magnitude of the hyperparameter. Moreover, if the hyperparameter updates are utilized less frequently, the learning rate should be higher.

In most experiments, we first measure the performance of the model trained using some fixed, random hyperparameters sampled from a reasonable interval. Next, we train the model with $T_1-T_2$ from that random hyperparameter initialization, measuring the final performance. Finally, we rerun the training procedure with the fixed hyperparameter set to the final hyperparameter values found by $T_1-T_2$. Note that in all the scatter plots, points with the same color indicate the same model configuration: same number of neurons and layers, learning rates, use of batch normalization, and the same types of hyperparameters tuned just with different initializations.

% % % % % % % % % % % % % % 
\subsection{Results}
% % % % % % % % % % % % % % 
\label{sec:results}

Figure \ref{fig:traject} illustrates resulting hyperparameters changes during $T_1-T_2$ training. To see how the $T_1-T_2$ method behaves, we visualized trajectories of hyperparameter values during training in the hyperparameter cost space. For each point in the two-dimensional hyperparameter space, we compute the corresponding test cost without $T_1-T_2$. In other words, the background of the figures corresponds to grid search on the two-dimensional hyperparameter interval. The initial regularization hyperparameter value is denoted with a star, while the final value is marked with a square.

As can be seen from the figure, all runs converge to a reasonable set of hyperparameters irrespective of the starting value, gradually moving to a point of lower log-likelihood. Note that because the optimal values of learning rates for each hyperparameter direction are unknown, hyperparameters will change the most along directions corresponding to either the local gradient or the higher relative learning rate.   

One way to use the proposed method is to tune the hyperparameters, and then rerun the training from the beginning using fixed values for the hyperparameters set to the final values acquired at the end of $T_1-T_2$ training. Figure ~\ref{fig:before} illustrates how much $T_1-T_2$ can improve initial hyperparameters. Each point in the grid corresponds to the test performance of a model fully trained with two different fixed hyperparameters: one is the initial hyperparameter before being tuned with $T_1-T_2$ (x-axis), the other is final hyperparameter found after tuning the initial hyperparameter with $T_1-T_2$ (y-axis). As can be seen from the plot, none of the models trained with hyperparameters found by $T_1-T_2$ performed poorly, regardless of how poor the performance was with the initial hyperparameters. 

\begin{figure*}
\centering
\includegraphics[width=0.32\textwidth]{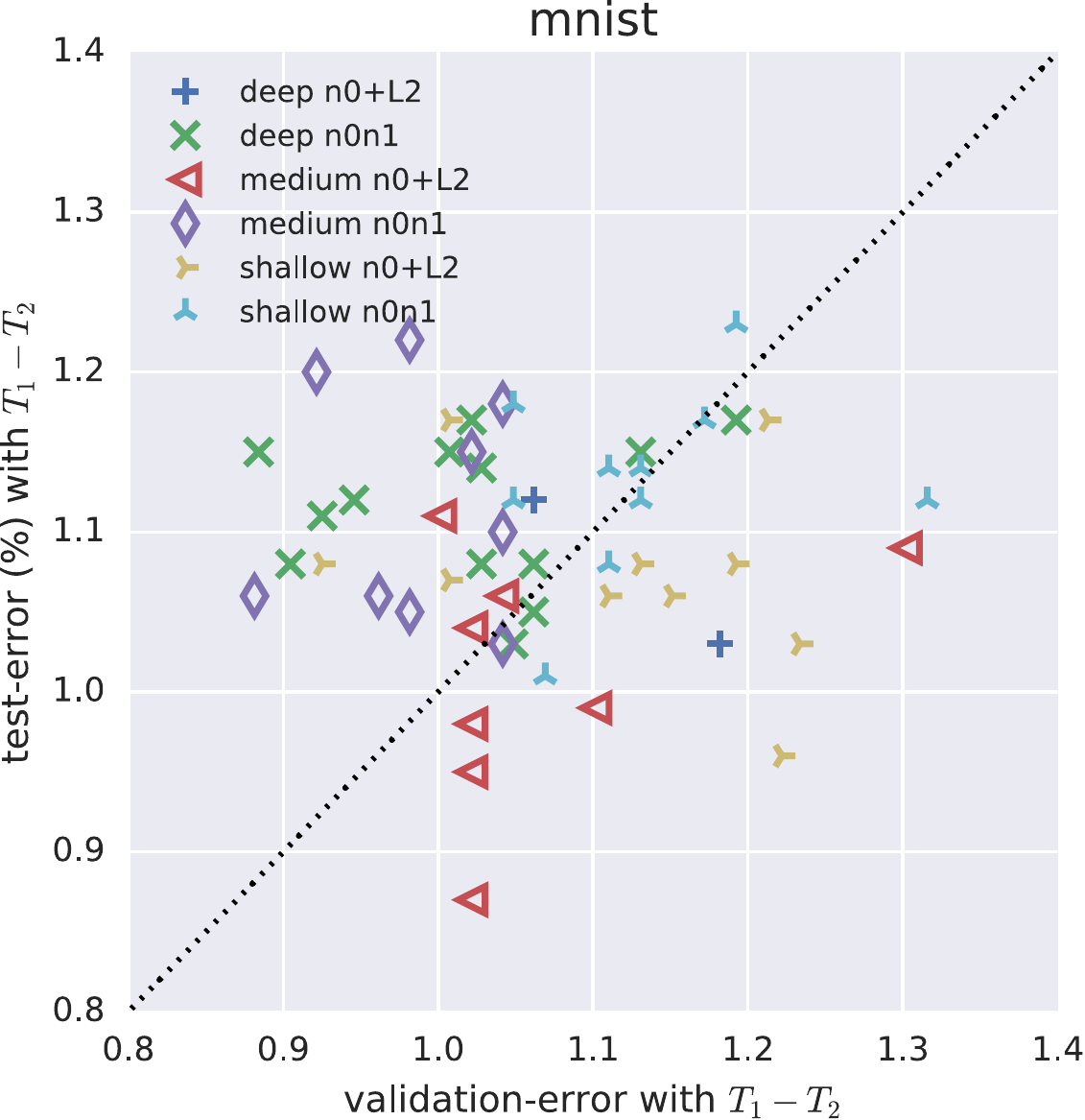}
\hfill
\includegraphics[width=0.32\textwidth]{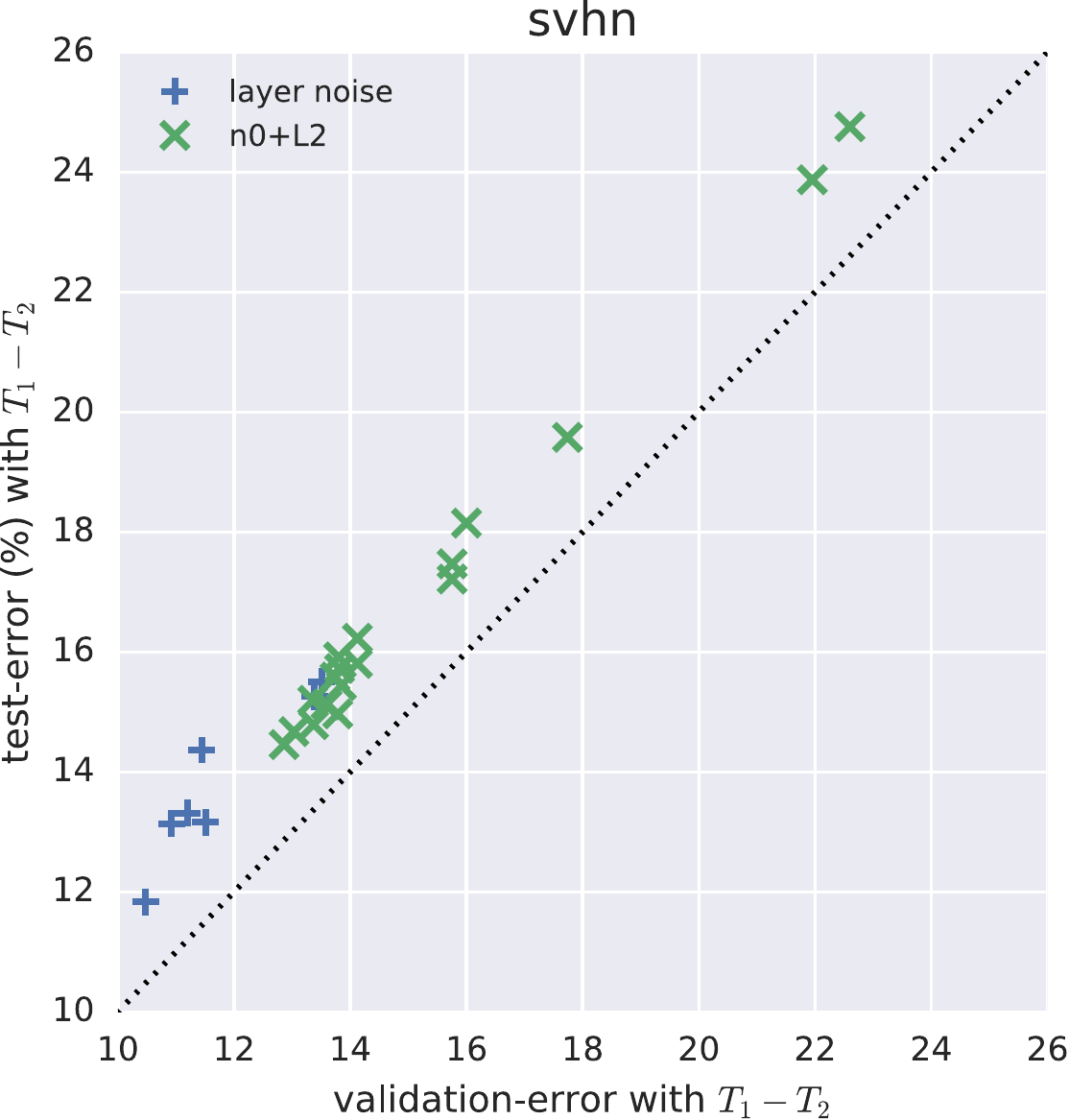}
\hfill
\includegraphics[width=0.32\textwidth]{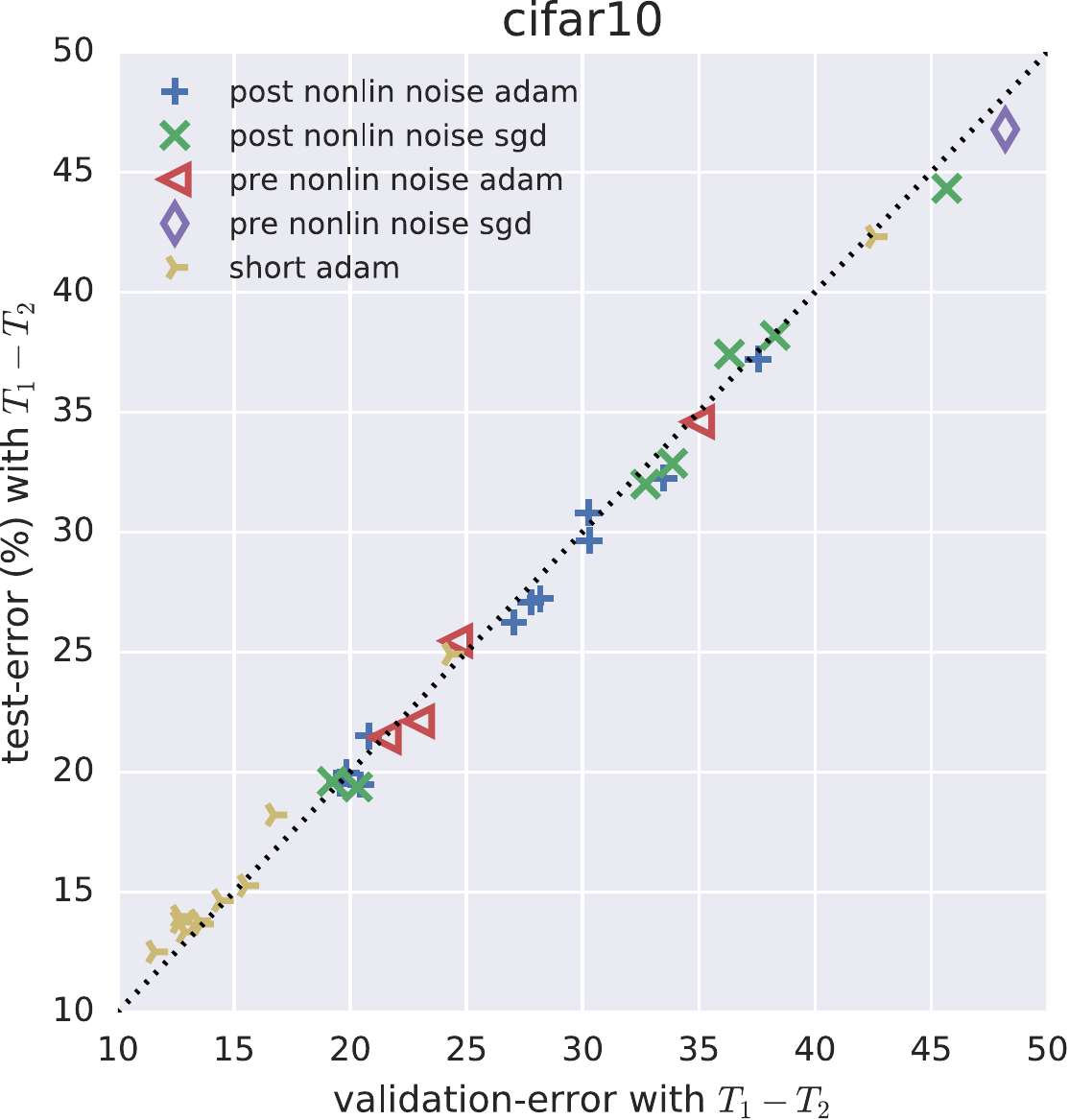}
\caption{Classification error of validation set vs test set, at the end of $T_1-T_2$ training for MNIST (left), SVHN (middle), and CIFAR-10 (right). 
For MNIST there is no apparent structure, but all the results lie in a region of low error. 
The results for the other two datasets correlate strongly, suggesting that validation set performance is still indicative of test set performance. 
%For SVHN we observe an offset which points to the test set being more difficult than the validation set.
}
\label{fig:q3}
\end{figure*}

\begin{figure*}[!ht]
\begin{center}
\centerline{\includegraphics[width=1.0\textwidth]{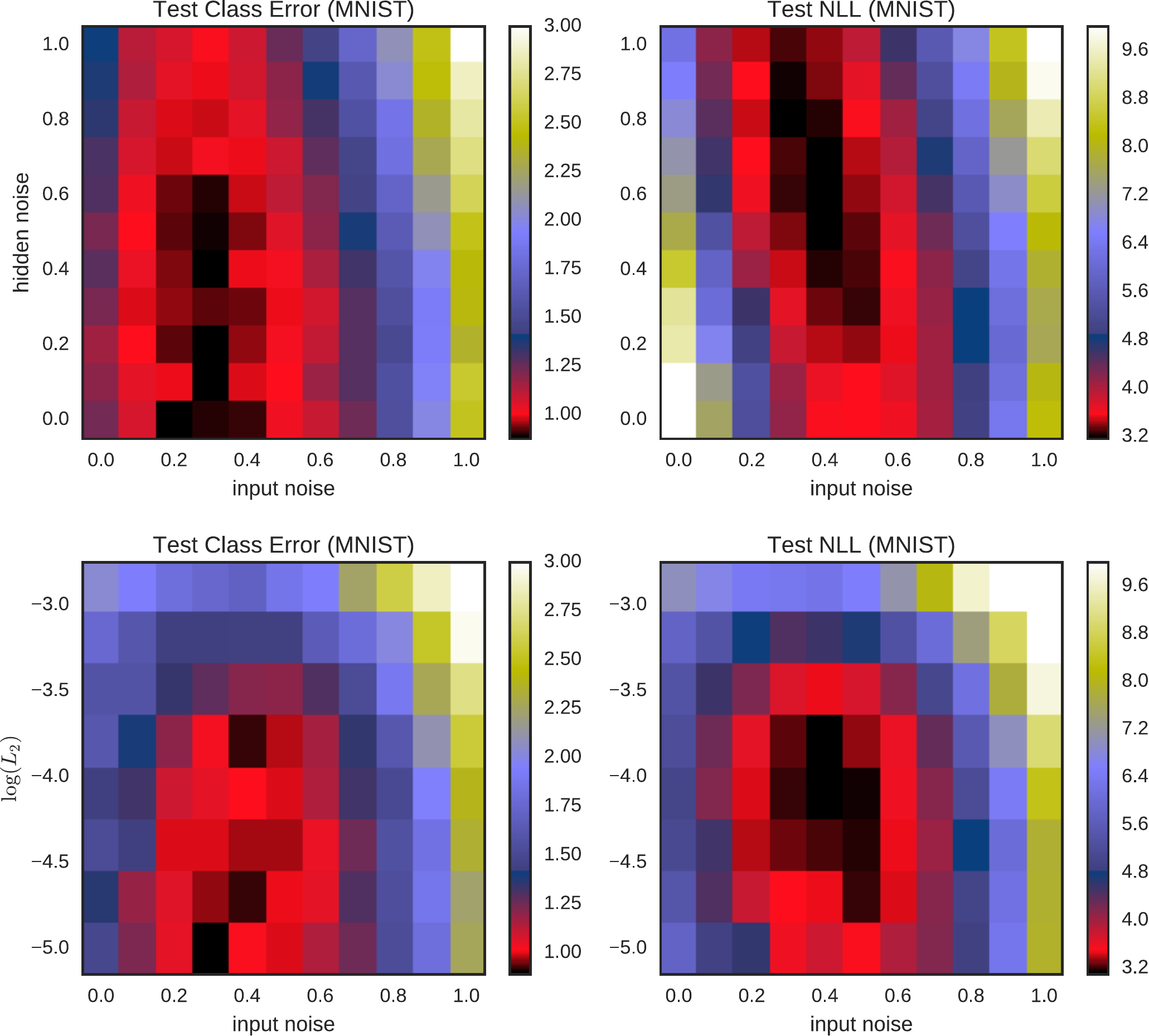}}
\caption{Grid search results on a pair of hyperparameters (no tuning with $T_1-T_2$). Figures on the right represent the test error at the end of training as a function of hyperparameters. Figures on the left represent the test log-likelihood at the end of training as a function of hyperparameters. 
We can see that the set of hyperparameters minimizing test log-likelihood is different from the set of hyperparameters minimizing test classification error.}
\label{fig:costerror}
\end{center}
\end{figure*}

Next we explore the strength of the hysteresis effect, i.e. how the performance of a model with a different hyperparameter history compares to the performance of a model with a fixed hyperparameter.
In \autoref{fig:q2} we plot the error after a run using $T_1-T_2$, compared to the test error if the model is rerun with the hyperparameters fixed to the values at the end of $T_1-T_2$ training. The results indicate that there is a strong correlation, with in most cases, reruns performing somewhat better. The method can be used for training models with fixed hyperparameters, or as a baseline for further hyperparameter finetuning. The hysteresis effect was stronger on CIFAR-10, where retraining produced significant improvements.

%\pagebreak

We explore the possibility of overfitting on the validation set. Figure~\ref{fig:q3} (right) shows the validation error compared to the final test error of a model trained with $T_1-T_2$. We do not observe overfitting, with validation performance being strongly indicative of the test set performance. For MNIST all the results cluster tightly in the region of low error, hence there is no apparent structure. It should be noted though, that in these experiments we had at most 20 hyperparameters, making overfitting to validation set unlikely.

\section{Discussion and Conclusion}

We have proposed a method called $T_1-T_2$ for gradient-based automatic tuning of continuous hyperparameters during training, based on the performance of the model on a separate validation set. We experimented on tuning regularization hyperparameters when training different model structures on the MNIST and SVHN datasets. The $T_1-T_2$ model consistently managed to improve on the initial levels of additive noise and L2 weight penalty. The improvement was most pronounced when the initial guess of the regularization hyperparameter values was far from the optimal value. 

Although $T_1-T_2$ is unlikely to find the best set of hyperparameters compared to an exhaustive search where the model is trained repeatedly with a large number of hyperparameter proposals, the property that it seems to find values fairly close to the optimum is useful e.g.\ in situations where the user does not have a prior knowledge on good intervals for regularization selection; or the time to explore the full hyperparameter space. The method could also be used in combination with random search, redirecting the random hyperparameter initializations to better regions.

\pagebreak

While the $T_1-T_2$ method is helpful for minimizing the objective function on the validation set, as illustrated in Figure \ref{fig:costerror}, a set of hyperparameters minimizing a continuous objective like cross-entropy, might not be optimal for the classification error. It may be worthwhile to try objective functions which approximate the classification error better, as well as trying the method on unsupervised objectives.

As a separate validation set is used for tuning of hyperparameters, it could be possible to overfit to the validation set. However, our experiments indicated that this effect is not practically significant in the settings tested in this paper, which had at most 10-20 hyperparameters.

The method could be used to tune a much larger number of hyperparameters than what was computationally feasible before. It could also be used to tune hyperparameters other than continuous regularization hyperparameters, using continuous versions of those hyperparameters. For example, consider the following implementation of a continuously parametrized number of layers: the final softmax layer takes input from all the hidden layers, however the contribution of each layer is weighted with a continuous function $a(i)$, such that one layer and its neighboring layers contribute the most,\ e.g. $output = softmax[\sum_{i=1..L} a(i) \tilde{W}_i \vect{h}_i ]$, where L is the number of layers and $a(i) = e^{(i-m)^2/v}$. Which layer contributes the most to the output layer, is determined with a differentiable function, and parameters of this function, $m$ and $v$ in this example, could in principle be trained using the $T1-T2$ method. 

\pagebreak

% % % % % % % % % % % % % % 
\section{Acknowledgements}
% % % % % % % % % % % % % % 
\label{sec:acknowledgements}

We are very thankful to many colleagues for the helpful conversations and feedback, particularly Dario Amodei and Harri Valpola. Special thanks to Antti Rasmus for the technical assistance. Also thanks to Olof Mogren and Mikael Kageback, who provided detailed comments on the paper draft. Jelena Luketina and Tapani Raiko were funded by the Academy of Finland. Mathias Berglund was funded by the HICT doctoral education network.

\bibliographystyle{natbib}
\bibliography{t1t2}

\end{document}